\documentclass[10pt,letterpaper]{article}
\usepackage[top=0.85in,left=2.75in,footskip=0.75in]{geometry}

\usepackage{amsmath,amssymb}

\usepackage{changepage}

\usepackage[utf8x]{inputenc}

\usepackage{textcomp,marvosym}

\usepackage{cite}

\usepackage{nameref,hyperref}

\usepackage[right]{lineno}

\usepackage[numbers]{natbib}

\usepackage{microtype}
\DisableLigatures[f]{encoding = *, family = * }

\usepackage[table]{xcolor}

\usepackage{array}

\newcolumntype{+}{!{\vrule width 2pt}}

\newlength\savedwidth

\raggedright
\usepackage[aboveskip=1pt,labelfont=bf,labelsep=period,justification=raggedright,singlelinecheck=off]{caption}

\makeatletter
\renewcommand{\@biblabel}[1]{\quad#1.}
\makeatother

\usepackage{lastpage,fancyhdr,graphicx}
\usepackage{epstopdf}
\fancyheadoffset[L]{2.25in}
\fancyfootoffset[L]{2.25in}
\newcommand{\AB}[1]{\textcolor{red}{#1}}
\begin{document}
\vspace*{0.2in}

\begin{flushleft}
{\Large
\textbf\newline{Unsupervised deep learning techniques for powdery mildew recognition based on multispectral imaging} 
}
\newline
\\
Alessandro Benfenati \textsuperscript{1},
Paola Causin \textsuperscript{2*},
Roberto Oberti \textsuperscript{3},
Giovanni Stefanello\textsuperscript{2}
\\
\bigskip
\textbf{1} Dept. of Environmental Science and Policy, Universit\`a degli Studi di Milano, Milano, Italy
\\
\textbf{2} Dept. of Mathematics, Universit\`a degli Studi di Milano, Milano, Italy
\\
\textbf{3} Dept. of Agricultural and Environmental Sciences - Production, Landscape, Agroenergy, Universit\`a degli Studi di Milano, Milano, Italy
\\
\bigskip

* paola.causin@unimi.it

\end{flushleft}
\section*{Abstract}

\subsection*{Objectives}
Sustainable management of plant diseases is an open challenge which has relevant  
economic and environmental impact.   
Optimal strategies rely on human expertise for field scouting under favourable conditions to assess the current presence and extent of disease symptoms. This labor-intensive task is complicated by the large field area to be scouted, combined with the millimeter-scale size of the early symptoms to be detected.
In view of this, image-based detection of early disease symptoms is an attractive 
approach to automate this process,  enabling a potential high throughput monitoring  at sustainable costs.

\subsection*{Methods}
Deep learning has been successfully applied in various domains to obtain an
automatic selection of the relevant image features by learning filters via a
training procedure. Deep learning has recently entered also the domain of plant disease
detection: following this idea, in this work we present a deep learning approach to
automatically recognize powdery mildew on cucumber leaves. We focus on unsupervised
deep learning techniques applied to multispectral imaging data and we propose the use
of autoencoder architectures to investigate two strategies for disease detection: i)
clusterization of features in a compressed space; ii) anomaly detection.

\subsection*{Results} 
The two proposed approaches have been assessed by quantitative indices. 
The clusterization approach is not fully capable by itself to provide accurate predictions but
it does cater relevant information. Anomaly detection has instead
a significant potential of
resolution which could be further exploited as a prior for supervised architectures
with a very limited number of labeled samples. 


\nolinenumbers

\section*{Introduction}

The accurate recognition and diagnosis of crop diseases
at early stages allows to apply crop-management systems based on timely, targeted and site--specific application of treatments. This enables a potential reduction of pesticide use and improved economic and ecological 
impact~\cite{cisternas2020systematic}.  Optimal strategies rely on human expertise for field scouting under favourable conditions to assess the current presence and extent of disease symptoms. This labor-intensive task is complicated by the large field area to be scouted, combined with the millimeter-scale size of the early symptoms to be detected.
The development of automated detection methods of crop diseases is thus an important  goal, which fits
 under the wide umbrella of the concept of precision agriculture. 
Investigations in this direction have included molecular analysis, spectroscopy, and analysis of volatile organic compounds but they are expensive and impractical to apply at commercial operating scales~\cite{martinelli2015advanced}.
In this respect, computer vision has an inherent great potentiality: symptoms of crop disease often cause a signature on plant leaves which can be detected by adequate strategies via   imaging-based techniques. Crop diseases have been detected and recognized by analyzing  color, texture, and shape of diseased leaves in images by different authors~\cite{gulhane2011detection,pixia2013recognition}. 
In addition, different spectral vegetation indices (VIs), {\em i.e.} algebraic combinations of reflectance values in two or more spectral channels, related to physiological
parameters have been proposed to differentiate healthy
from diseased plants~\cite{delalieux2008near, vigier2004narrowband}.  
In these approaches the selection of the relevant features
is however still dependent on the intervention of human experts. 
Recent developments in machine learning have opened new possibilities, paving the way to exploit data from optical sensors in crop disease detection with an automatic recognition of relevant features~\cite{zhang2017leaf}.
Specifically, deep learning (DL) methods based on convolutional neural networks (CNNs) have proven to produce 
accurate results. 
Studies based on leaf images obtained with conventional RGB cameras have proven the potentiality of this approach.
For example, Mohanty~\cite{mohanty2016using} used large CNNs  (GoogleNet and Alexnet) to classify 26 diseases 
over 14 crop species. A dataset consisting
of 54,306 labeled color images from the PlantVillage repository was considered for
the training phase. Sladojevic~\cite{sladojevic2016deep} used the CaffeNet CNN to 
classify 13 diseases over various crop plants. Training of the net was carried out with 4483 (augmented to 30,000) images downloaded from the web
and submitted to preliminary human screening filter and labeling.  Transfer learning techniques were adopted 
in both the above references to specialize CNNs to the application at hand. Other works further contributed on this line.
The authors  in~\cite{fuentes2017robust} considered 5000 images of tomato leaves with manually annotated bounding boxes  containing disease spots to
train various Region Proposal Networks to generate object
proposals. They successively channeled these attention areas into ``deep feature extractors'' such as the VGG net or a Residual Network to obtain accurate disease classification under diverse field conditions.
The authors in~\cite{wspanialy2020detection} used 16415 diseased tomato leaf images and 1590 healthy tomato leaves, upon classification of the disease by experienced observers.  The proposed algorithm utilizes a Res-Net architecture 
to classify leaves into healthy or diseased and a U-net architecture to semantically segment a subset of the images
to evaluate the severity of the disease.  
Semantic segmentation was performed as well in~\cite{lin2019deep} to recognize powdery mildew spots
on cucumber leaves, using a U-net architecture trained with 30 annotated samples (augmented to 10,000). 
All the previously mentioned approaches do automatically detect the relevant features, provided that a substantial amount of work is initially done by an human expert to  label/annotate the images. Even when transfer learning techniques are used, and thus   
already available labels may be taken advantage of, further labeling for the specific case is however required. 

In this work, we aim to leverage the capabilities of DL methods to achieve a preliminary but fairly 
accurate automatic detection of 
plant diseases, including early stage conditions, using unsupervised techniques. 
A similar concept, albeit obtained with different DL approaches, was
pursued in~\cite{behmann2014detection}, where the authors combined 
an unsupervised method ($k$-Means clustering) for extracting labels at pixel scale and a supervised method for classification (Support Vector Machine) at pixel and plant scales
for the detection of early stages of drought stress. 
Here we consider the detection of powdery mildew on cucumber ({\em Cucumis sativus}). Powdery mildew is a major foliar disease caused by different fungi in many crops  (vegetables, fruits, cereals etc) with common symptoms: the proliferation of hyphae filaments of the mycelium on the hosting tissue affects leaf reflectance to incident light,  leading to a whitish-gray, powdery appearance. At 
early to middle stages of infection, these thin filamentous structures have still low influence on the spectral signature of the leaf surface due to their small dimensions, 
low density, and spatial arrangement. This makes the early detectability of the disease a non--trivial problem. 
We use images obtained from proximal multispectral sensing,    
exploiting the altered spectral signature of diseased leaves not only in the visible (RGB) spectrum but also in the near-infrared (NIR)
band~\cite{mahlein2016plant, lowe2017hyperspectral,saleem2019plant}. 
We propose two strategies for disease detection: a clustering approach and, as a preferred choice,  
an unsupervised automatized feature extraction approach based on an anomaly score produced by an autoencoder (AE) neural architecture. Leaves whose anomaly score exceeds a certain threshold are considered diseased.



\section*{Materials and methods}

\subsection*{Background}
\label{sec:backg}

Leaf reflectance features have a high potential in detecting deviations from the healthy status of plants linked to dysfunction of the photo-system or destruction of the photo-chemical pigments, modifications in plant tissue composition and structure, or to the development of pathogen spores or propagules on the leaf surface. These biophysical modifications induce significant changes in the spectral signature of plant tissue that can be detected with adequate techniques~\cite{west2003potential,sankaran2010review}. 
Changes in the visible (VIS, 400 to 700~nm) and near-infrared (NIR, 700 to 1100~nm) spectral ranges are of particular relevance since they can be measured with common silicon-based sensors or cameras. 
In these bands, healthy leaves typically exhibit (a) low reflectance at VIS wavelengths owing to strong absorption by pigments; (b) high reflectance in the NIR owing to internal scattering in the leaf structure, except for weak water absorption at specific wavebands.
General disease symptoms correspond to discrete structures or lesions on leaf tissue evolving from millimeter-scale size to macroscopic patches, and are characterised by an increased reflectance in VIS range, especially in the chlorophyll absorption bands in the blue (430-470~nm) and red (630-690~nm) bands. Conversely, at more advanced stages of disease, reflectance in NIR range on symptomatic areas is reduced by oxidation and senescence processes in the tissue, and at plant canopy scale by decreased biomass growth, defoliation and drying.
These general features hold for the specific case of powdery mildew and upon this it relies the rationale of using multi-spectral imaging in the above indicated bands to detect regions in leaf surface exhibiting deviations from healthy spectral signatures.

\subsection*{Plant material and disease inoculation}
\label{sec:vasi}

Plants of cucumber ({\em Cucumis sativus}) were sown and grown in pots under controlled conditions in greenhouse at 25/22$^\circ$C (day/night),
60\% relative humidity. Plants were regularly watered and fertilized as needed, and no pesticide treatment was applied. At a development stage of 3 leaves, a group of plants was separately inoculated with isolates of {\em Podosphaera xanthii} by spraying a suspension of freshly sporulating colonies onto leaves. 
The rest of the plants were kept isolated under controlled conditions in order to maintain healthy conditions during the growth.
Multiple lots of plants were subsequently cultivated and inoculated to provide enough samples to the aim of the experiment.

\subsection*{Multispectral images acquisition and preprocessing}
\label{sec:images}

In order to obtain a wide range of powdery mildew symptoms, the inoculated plants were sampled at different dates, {\em i.e.} after 5, 10, 15 days from the inoculation, and imaged together with age-companion healthy plants. Healthy and diseased cucumber leaves were imaged via 
a QSi640 ws-Multispectral camera (Atik Cameras, UK) equipped with a Kodak 4.2~Mp micro-lens image sensor and 8 passband spectral filters operating at wavebands from 430 to 740~nm. 
For the purpose of this experiment, leaves were imaged singularly on a dark background, under controlled diffuse illumination conditions. 
Images were acquired in the single spectral channels 
430~nm (blue, B), 530~nm (green, G), 685~nm (red, R) and 740~nm (near--infrared, NIR). 
A set of RGB images of the same leaves in standard CIE color space
were also acquired for reference.  Camera parameters were set and 
image collection was performed via an {\em in--house} 
developed acquisition software written in MATLAB. 
Reflectance calibration of the grey-level intensity of the pixels at different acquisitions was carried out by including in each image 3 reflectance references targets (Spectralon R = 0.02, R = 0.50 and R = 0.99; Labsphere, USA).
We obtained two datasets: a first dataset, named Dataset~A, consisting in~64 pictures 
 of healthy leaves and 70 pictures of diseased leaves with mild to severe symptoms and a second dataset, named Dataset~B, consisting in 33 pictures of healthy leaves and 44 pictures of diseased leaves with mild symptoms. The complete dataset was thus composed of 97 healthy samples and 114 diseased samples.
Starting from an original resolution of 2048$\times$2048, each image was cropped and resized in order to get the resolution down to 512$\times$512, which was more manageable.  We preprocessed the dataset with min-max normalization in order to obtain values within the interval~$[0,1]$. Using the NIR channel, where leaves are highly distinguished from the background, we
also  computed binary masks that indicate which pixels belong to the leaves and which belong to the background. 
 \begin{figure}[ht]
\centering
\includegraphics[width=.7\linewidth]{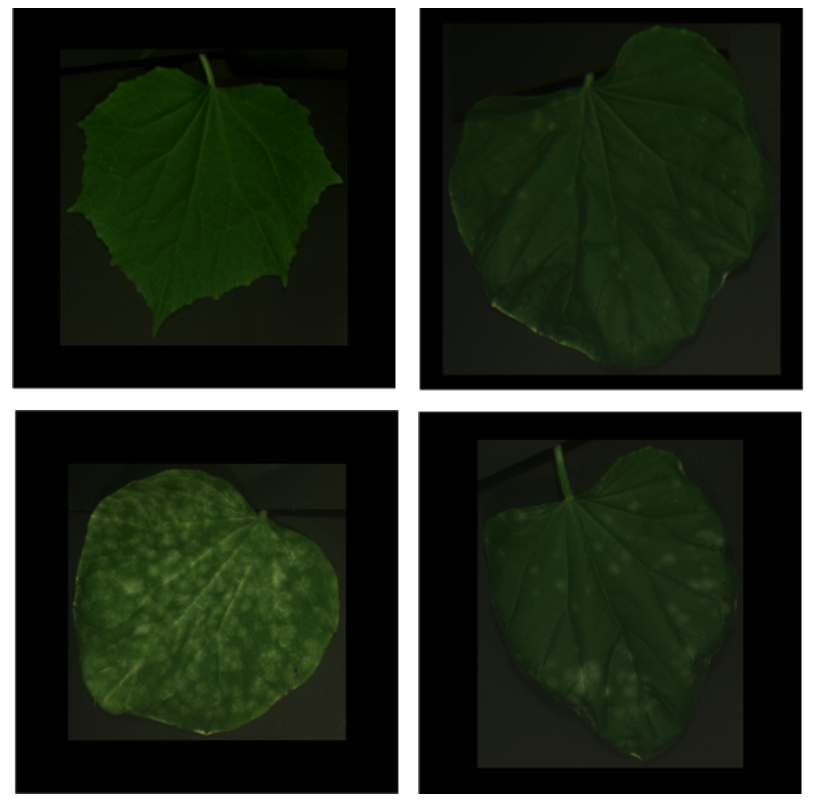}
\caption{{\bf  RGB images of four samples of cucumber leaves from the acquired dataset.} The leaves show different degrees 
of powdery mildew, from very mild (top left) to severe, in clockwise order.}
\label{fig:samples}
\end{figure}

\subsection*{Deep anomaly detection}

\subsubsection*{Automatic feature extraction via DL networks}
Given a dataset $\mathcal{X} = \{x_1, x_2, \dots , x_N \}$ with $x_i \in R^D$, $i=1,\dots, N$,
 let $\mathcal{Z} = \{z_1, z_2, \dots , z_K \}$, with $z_i \in R^D$, $i=1,\dots, K$
 with $K \ll N$, be a representation
space of $\mathcal{X}$.  We aim at learning a mapping function
$\phi(\cdot): \mathcal{X} \rightarrow \mathcal{Z}$ such that 
$$
z=\phi(x;\Theta),
$$
in a such a way that anomalies (diseased leaves) can be
easily differentiated from normal (healthy leaves) data instances in the compressed 
space yielded by the 
mapping~$\phi$. In our context, $\phi$ is a neural network--enabled mapping function with
learnable parameters~$\Theta$.
Specifically, we use convolutional autoencoders
to learn a low-dimensional feature representation space 
on which the given data instances can be well reconstructed. 
An autoencoder (AE) is composed of encoder and decoder blocks: the encoder maps the original data onto the low-dimensional feature space, while the decoder attempts to recover the data from the projected low-dimension feature space. Training is performed by minimizing the distance between
the original data and their decoded version.
A basic AE formulation reads:
\begin{equation}
\begin{array}{l}
z=\phi_e(x;\Theta_e), \qquad \widehat{x}=\phi_d(z;\Theta_d), \\[3mm]
\{\Theta_e^*,\Theta_d^*\}=\underset{\{\Theta_e,\Theta_d\}}{\arg\min}\displaystyle\sum_{x \in \mathcal{X}}
||x-\widehat{x}||^2, \\[5mm]
\end{array}
\label{eq:basicF}
\end{equation}
where $\phi_e$ is the encoder part
of the network with parameters $\Theta_e$ and $\phi_d$ is the decoder part
of the network with parameters $\Theta_d$. 
The 
learnable parameters are the weights of the convolutional filters.

\subsubsection*{Feature extraction and clustering}

Cluster analysis is the process of finding “natural” groupings by gathering “similar”
objects together, according to some similarity measure. 
Clustering is  a notoriously hard task, whose outcome
is affected by a number of factors – among which data dimensionality.
As a matter of fact, one must face the problem that not all the
original features are relevant for clustering and high
dimensional data may lead to algorithm break--down. 
The use of preprocessing strategies such as dimensionality reduction 
allows clustering to perform better. 
In this work, we use the AE framework to extract 
a feature set on which to carry out clustering. 
To perform this latter task, we use a classic $k$-means approach and we seek cluster centroids by an iterative 
optimization process that minimizes the Euclidean distance between data points in the feature space
produced in the bottleneck of the AE and their nearest centroid. 
More sophisticated approaches have been proposed in literature, for example using clustering
as a prior for image classification, iteratively exploiting the clusters to enrich the training dataset~\cite{piernik2021study}. 
In this study, we limit ourselves to use clustering to prove the discriminating potential
embedded in the feature representation space and we highlight what are in our
experience the potentialities of this approach.

\subsubsection*{Feature extraction and anomaly detection}

Anomaly detection, also known as novelty detection, is 
the process of detecting data instances that deviate from a given set of samples. 
Anomaly detection can be carried out via a neural network by training  
the net on normal samples so to build a feature representation of ``normality''.
 An anomaly score is introduced which quantifies the discrepancy of a sample
 from its reconstruction performed by the net. In this framework, normal samples will have a low anomaly score,
 based on a given threshold, while if a sample presents anomalies, it will yield a higher anomaly score.
We use the same framework of~(\ref{eq:basicF}) 
and we endow it with the score 
\begin{equation}
s_x=\frac{\|x-\widehat{x}\|^2}{{\|x\|^2}},
\label{eq:score}
\end{equation}
where $\widehat{x} =\phi_d(\phi_e(x;\Theta_e^*);\Theta_d^*)$.
If, for an instance $\widetilde{x}$, it happens that $s_{\widetilde{x}} \ge \Gamma$,
$\Gamma$ being a set threshold, 
then $\widetilde{x}$ is classified as an anomaly. 
One can also use the anomaly score 
based on the discrepancy between the compressed representation  
$z =\phi_e(\widetilde{x}; \Theta_e^*)$ of the sample $\widetilde{x}$ 
and the compressed representation of its reconstruction $\widetilde{z} =\phi_e(\widehat{\widetilde{x}}; \Theta_e^*)$,
{\em i.e.},
\begin{equation}
s_z=\frac{\|z-\widetilde{z}\|^2}{{\|z\|^2}}.
\label{eq:score}
\end{equation}

\section*{Results and discussion}

\subsection*{Clustering approach}

\subsubsection*{Neural network architecture and implementation}
\begin{figure}[ht]
\centering
\includegraphics[width=\linewidth]{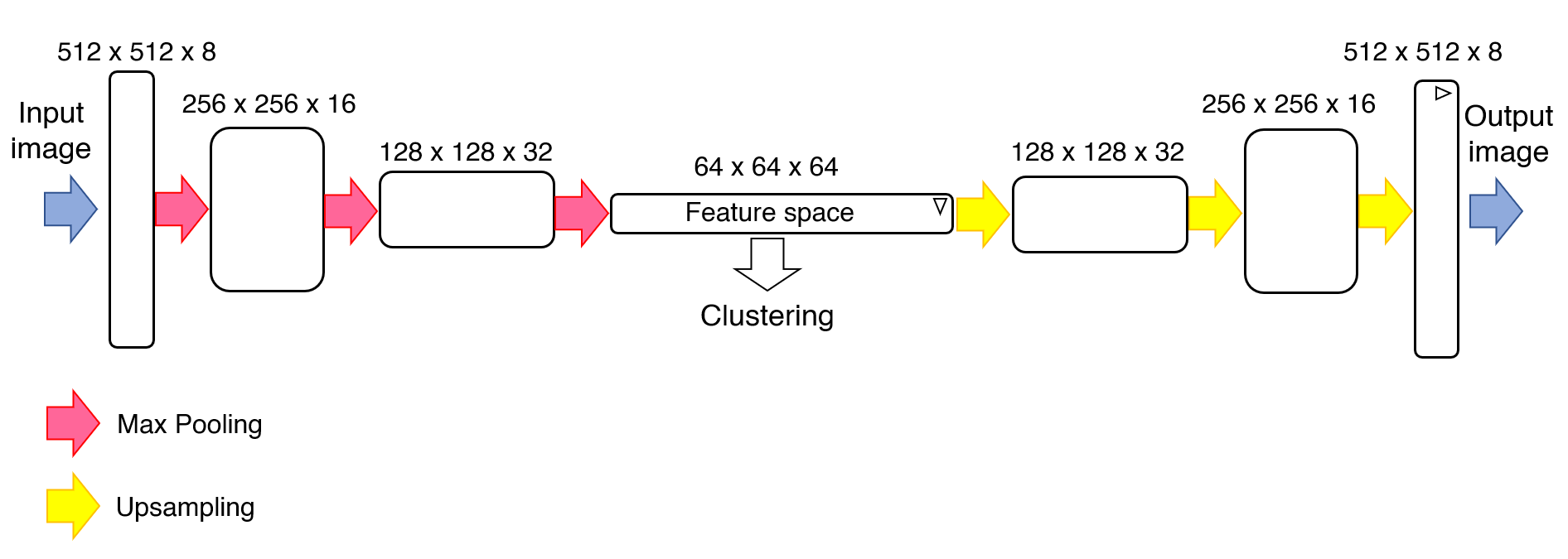}
\caption{{\bf  Structure of the convolutional autoencoder used in the clustering approach (Clu-AE model).}
Each block is composed of a convolutional layer, a batch normalization layer and a ReLU activation. The block marked with the symbol $\triangledown$ has a dropout layer at its end. The block marked with 
the symbol $\triangleright$ is followed by a $1\times 1$ kernel convolutional layer with a number of features equal to the number of channels of the input image.}
\label{fig:AECluster}
\end{figure}

We define the model as follows. 
The encoder part is composed of 4  blocks, each made of a convolutional layer, a batch normalization layer and a ReLU activation layer. The blocks are connected through max pooling layers in order to decrease the resolution of the image. The number of features for each block is 8,16,32,64, going from the shallowest block
to the deeper. At the lowest level, a dropout layer is inserted to reduce over fitting and training time. In our experiments, we found that the drop--out also positively encouraged the model to learn diverse, 
non--redundant features. The decoder part mirrors the encoder, without the last block: it is composed of 3 blocks, with 32,16,8 filters, connected through upsampling layers in order to increase the resolution of the image back to its original size. After the last decoder block, a convolutional layer with a $1 \times 1$ kernel and a number of filter equal to the channels of the original image coupled with a logistic activation function outputs the reconstruction of the input image (see Fig~\ref{fig:AECluster}). For brevity, we shall name in the following this structure as Clu-AE.

For this study, we used the R,G,B, and NIR channels and we also added a VI channel which consists in the ratio between the NIR channel and the R channel (NIR/R). 
The dataset has been split in training, validation and test as follows: 5\% of the whole dataset has been used for testing,  5\% of the remaining samples has been used for validation and the remainder has been used to train the model. 
{The training was performed via the Adam optimizer with a learning rate $\eta = 10^{-3}$, default hyperparameters and initialization of the parameters via the He strategy.}
The maximum number of epochs to train the model was set to 500 and early stopping was implemented, monitoring 
the validation loss with a patience parameter of 20 epochs. The batch size was 
set to 8. After each epoch, the whole training set was shuffled. As for clustering, the
$k$-means algorithm was executed 20 different times with randomly initialized centroids for each experiment, and the partition that achieved the best results is kept  (incidentally, this
partition was the one that the majority of the runs converged to).

\subsubsection*{Evaluation metrics}
In order to evaluate the clustering quality, several different metrics have been proposed. Here, we consider:
\begin{itemize}
\item  Silhouette coefficient \AB{\cite{ROUSSEEUW198753}}, defined as:
$$
S(i) =\frac{d_s(i) −d_a(i)}{\max\{d_a(i),d_s (i)\}}
$$
where $d_a(i)$ is the average distance of point $i$ from all other points in its cluster and $d_s(i)$ is the smallest average distance of $i$ to all points in any other cluster. The Silhouette coefficient measure how well each individual point fits in its cluster: if $S \simeq 0$, the point is right at the inflection point between two clusters; if $S \simeq -1$  the point would be better assigned to  another
cluster, if if $S \simeq 1$, the point is well-assigned to its cluster.
For an evaluation of the clustering quality at a global level, instead of a point-wise level, it is common to average the Silhouette coefficients of all the points to give the Average Silhouette coefficient (aSC);
\item Davies-Bouldin index (DB) \AB{\cite{4766909}}, defined as 
$$
DB=\frac{1}{n}\sum_{i=1}^n \max_{j \ne i}\frac{\sigma_i+\sigma_j}{d(c_i,c_j)}
$$
 where $n$ is the number of clusters,  $d(c_i,c_j)$ is the distance between the centroid of cluster $i$ and
 cluster $j$, and $\sigma_i$ is the average distance of all points in
 cluster $i$ from its centroid $c_i$. The DB index leverages the concept that 
 very dense and well spaced clusters constitute a good clustering. The minimum score is zero, and differently from most performance metrics, the lower the value, the better the clustering performance.
\end{itemize}

\subsubsection*{Reconstruction and compression}
\begin{figure}[ht]
\centering
\includegraphics[width=\linewidth]{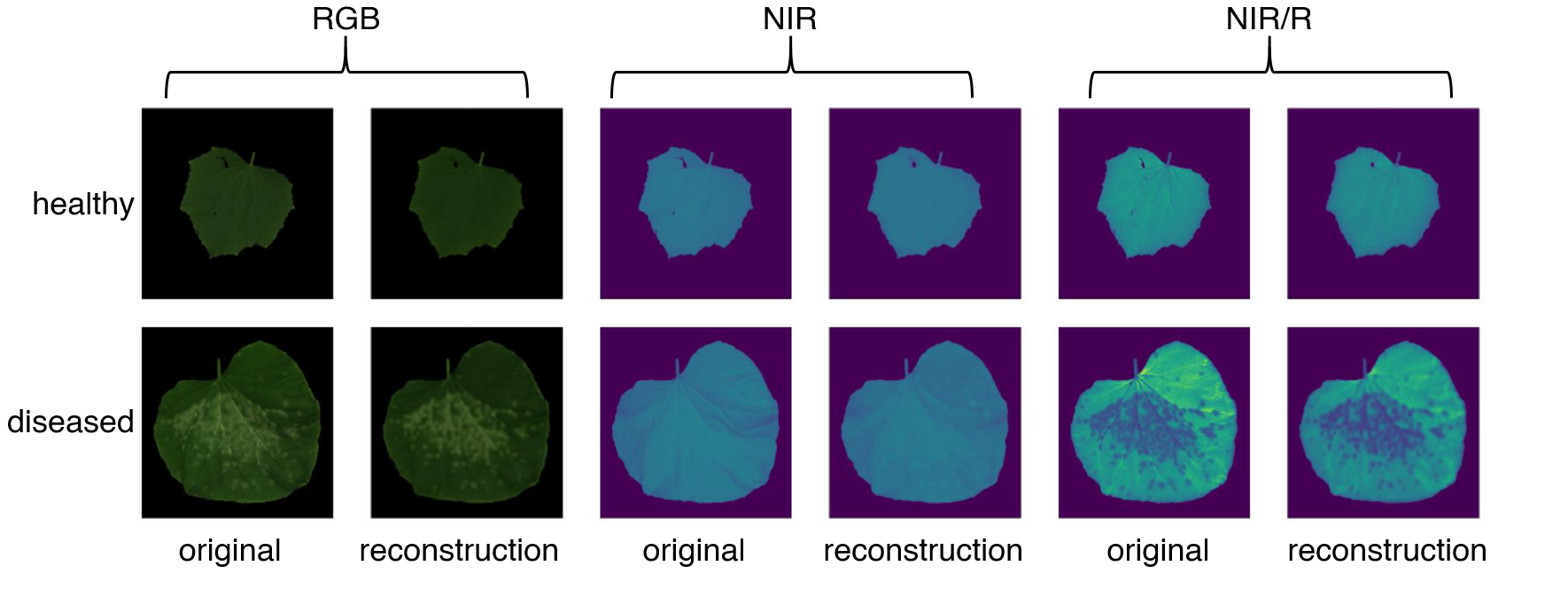}
\caption{{\bf  Clu-AE model: comparison between original and reconstructed images of two random leaves.}
 For each channels, RGB, NIR, and NIR/R, the original channel is shown 
 on the left and its reconstruction on the right. The first row refers to a healthy leaf, the  second row to a diseased leaf.}
\label{fig:comparison1}
\end{figure}

\begin{figure}[ht]
\centering
\includegraphics[width=\linewidth]{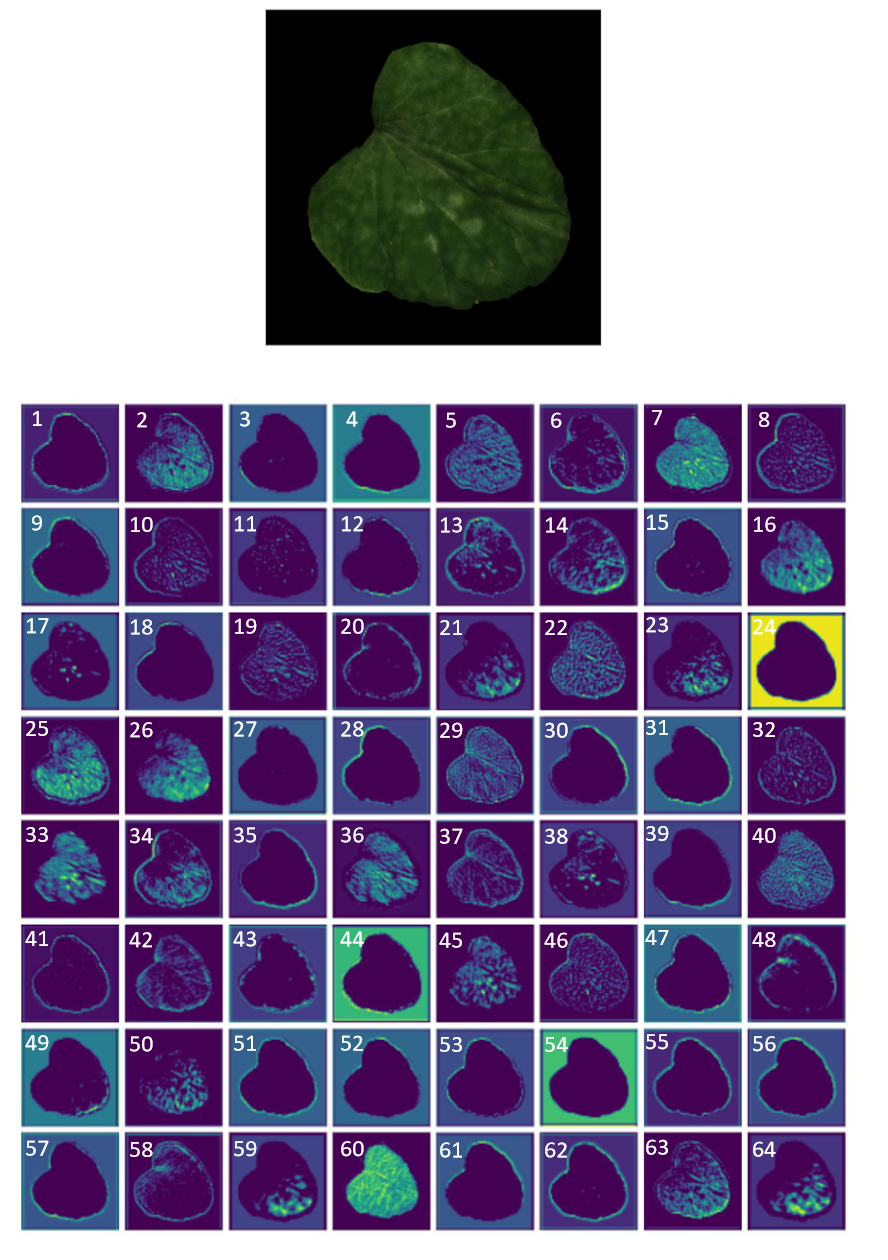}
\caption{{\bf  Visualization of the representation space extracted by the Clu-AE model.} The input
image is the (diseased) leaf shown in RGB. Each of the 64 features has 
its own range of values; the brighter the color of a pixel in the feature image, the higher its numerical value.}
\label{fig:feature1}
\end{figure}

We start by checking the reconstructive power  of the Clu-AE model.
In Fig~\ref{fig:comparison1} we show for one random healthy leaf
and one random diseased leaf the original datum~$x$ and the reconstructed
datum~$\widehat{x}$.  
The autoencoder is able to reconstruct the leaves, together with characteristic attributes like veins or other spots, with a satisfactory accuracy. A certain degree of blurriness is however present in the reconstructed images: {this is a common issue when AEs are employed in imaging processing}.
In Fig~\ref{fig:feature1},  we visualize the learned compressed feature map (64 images) for a 
random diseased leaf. It is evident that many of the features focus more on the shape of the leaf rather than on its inner part, by encoding the leaf as a black shape and highlighting its edge, distinguished from the background
(for example in features: 1, 3, 4, 9, 15, 28, 30, 31, 51, 52, 61, 62). Those features are not all identical but each one lights up on different portions of the leaf edge. Other features encode the information in the interior of the leaf, highlighting leaf veins (for example, 
in features: 2, 29, 37), different shades of healthy tissue (for example, in features: 7, 29, 37, 60) or the presence and degree of severity of disease spots (for example, features: 13, 32, 33).
Not all the features are easily interpretable, as some of them light up in ways that do not have a clear meaning for a human observer.

\subsubsection*{Clustering results}
The aim of clustering is to identify and separate the sets of healthy and diseased leaves.
These latter show different degree of severity,
ranging from mild to necrotic. 
We report in Tab~\ref{tab:clu1}  the evaluation metrics of the clustering performance when all of the 64 extracted features are considered.  The number of clusters is considered here as 
an hyperparameter to be chosen.
\begin{table}[bth]
\centering
\begin{tabular}{|l|c|c|}
\hline
           & aSC                            & DB                             \\ \hline
2 clusters & {\bf 0.17056} & 2.33102                        \\\hline
3 clusters & 0.13617                        & 2.21739                        \\\hline
4 clusters & 0.14190                        & {\bf 2.14507}\\ \hline
\end{tabular}
\caption{Evaluation metrics for the Clu-AE approach obtained using all the 64 extracted features. In bold
the best performance for each index.}
\label{tab:clu1}
\end{table}

\begin{figure}[htbp]
\centering
\includegraphics[width=.8\linewidth]{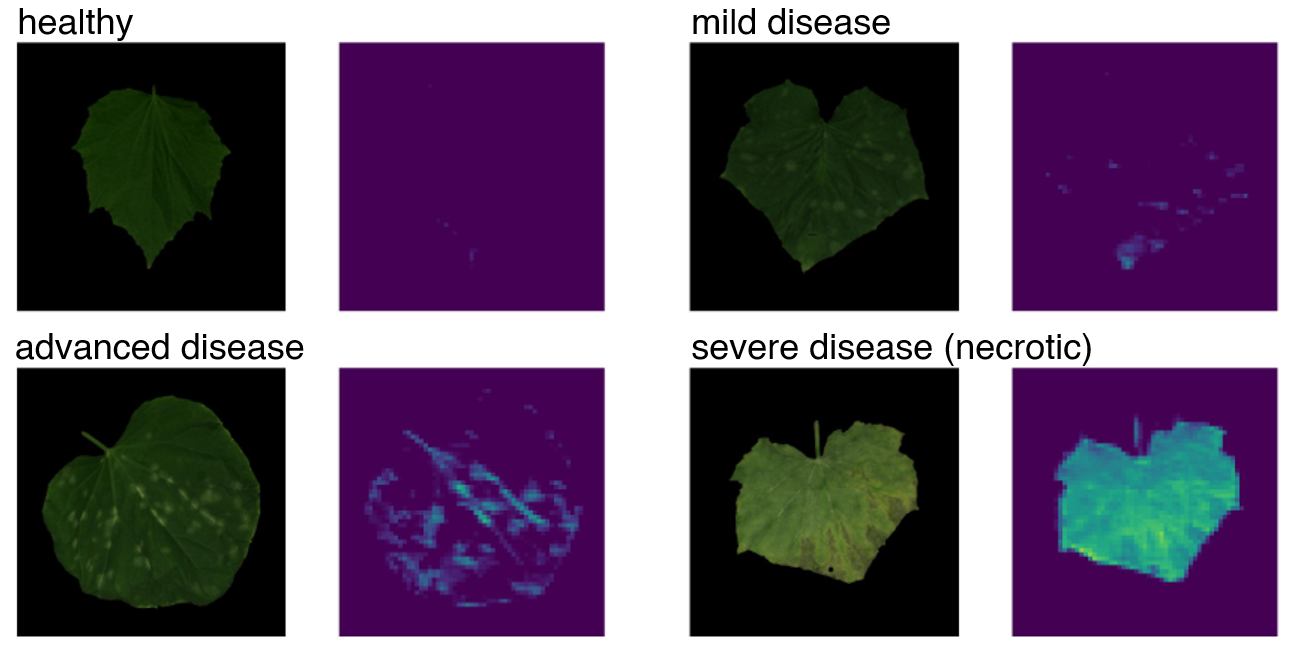}
\caption{{\bf  Example of  responses of feature no.33 to different input leaves.} 
The feature lights up in correspondence of disease spots 
(values are normalized considering all the leaves in the dataset). }
\label{fig:feature2}
\end{figure}

The best value for each index is attained by a different number of clusters, 
as the aSC is maximum for 2 clusters and the DB index is minimum for 4 clusters.
When considering 2 clusters, the first cluster contains the majority of the samples, where both healthy and diseased leaves are grouped together, even including almost all the most severely affected leaves. The other cluster contains again both healthy and diseased leaves, where it is arguable that many of the leaves are similar in terms of their general shape and dimension. When considering 4 clusters, the first and third clusters contain both healthy and diseased leaves, while the fourth cluster contains only healthy leaves. The second cluster is instead composed of the most severely diseased leaves. This 4--cluster partition fails when trying to establish if a certain leaf with no evident signs of infection is diseased or not. In general, we lose the dichotomy “healthy cluster” vs “diseased cluster”, but it seems that the clusters are now dictated mainly by dimension and shape of the leaves rather than their health condition, which is especially evident in the first, third and last cluster. We believe that the reasons behind this failure are mainly two: first, the dimension of the data, even if it was reduced by 80\% in the encoding, is still too high for the $k$-means algorithm. Second, too many features detect only the edge of the leaf, which explains the tendency of the model to partition the leaves based on their shape and dimension. 
To improve these results, we tried to cluster a compressed version of our data based only on
a restricted number of features which we deemed more relevant for our objective. 
After multiple attempts, we obtained the best results by selecting ({\em cherry-picking}) only 
one feature, no.33 (see Fig~\ref{fig:feature1}).  In Fig~\ref{fig:feature2}, we report some of the responses of this feature 
for input leaves that present different levels of infection.  It is clear that this feature is responsive to disease spots, the more severe the disease spot, the higher the response.
We report in Tab~\ref{tab:clu2} the evaluation metrics of the performance when only feature no.33 is considered. Both the evaluation metrics in this case have improved considerably compared to the previous case. This time, both the aSC and DB index attain their best values for the 2 clusters partition. One of the reasons for this improvement is that selecting only one feature out of 64 greatly reduces the dimension and clustering performs better.
However, even if the metrics now attain better values, the clusters are still not really informative. 
In the case of 2 clusters, the first cluster contains the majority of the samples, where both healthy and mildly diseased leaves are grouped together, while the other one mainly contains the most extremely diseased leaves.  This may be due to the fact that the extracted feature is by itself not informative enough when it comes to detecting all the diseased spots: for example,  it does not light up in certain particularly faint spots. On the other hand, mildly 
diseased leaves appear much more similar to healthy leaves than to diseased ones at the feature level, and for this reason they tend to get grouped together with healthy leaves. This problem is not solvable by increasing the 
granularity (number) of clusters, as experiments have shown that the clusters still group healthy and diseased leaves together, and are difficult to interpret. 
\begin{table}[bht]
\centering
\begin{tabular}{|l|c|c|}
\hline
           & aSC                            & DB                             \\ \hline
2 clusters & {\bf 0.58295} & {\bf 0.91016}                        \\\hline
3 clusters & 0.33386                        & 1.38958                        \\\hline
4 clusters & 0.32163                        & 1.87777\\ \hline
\end{tabular}
\caption{Evaluation metrics of the Clu-AE approach when only feature no.33 of Fig~\ref{fig:feature1} is used. In bold
the best performance for each index.}
\label{tab:clu2}
\end{table}

\subsection*{Anomaly detection approach}

\subsubsection*{Neural network architecture and implementation}

\begin{figure}[h]
\centering
\includegraphics[width=\linewidth]{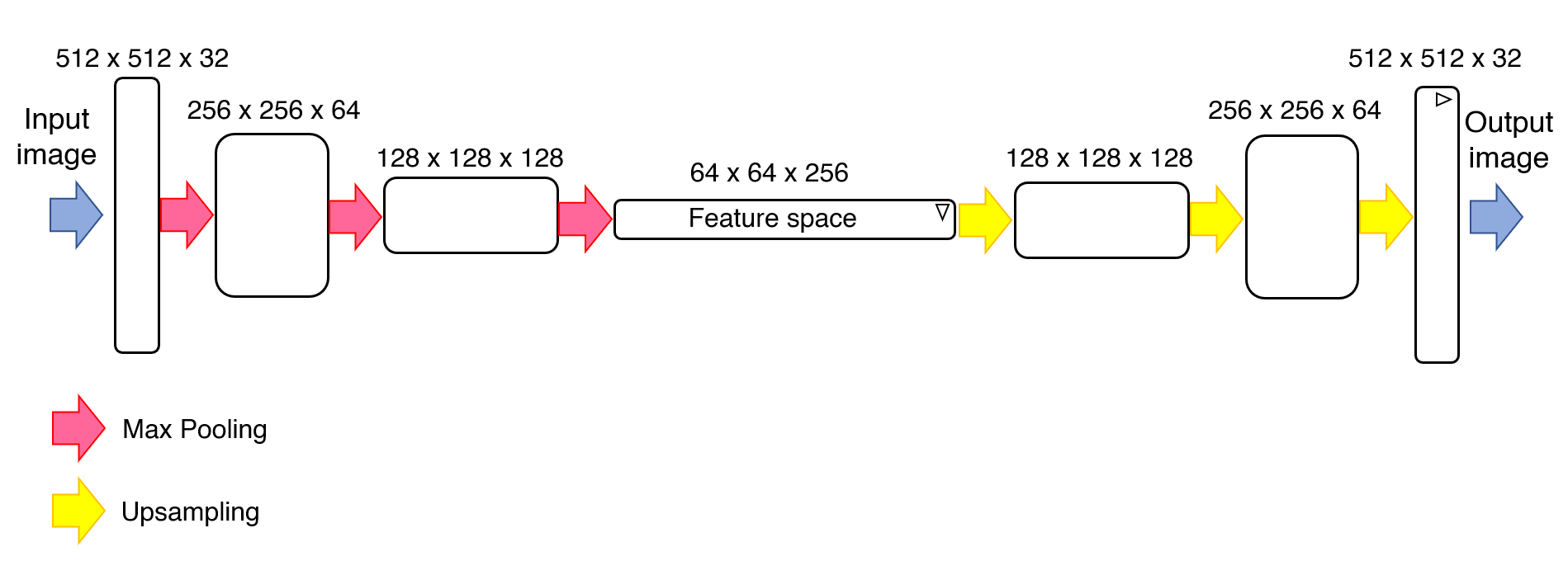}
\caption{{\bf  Structure of the convolutional autoencoder used in the anomaly detection algorithm (filter sizes 
correspond to the Ano-AE model).}
Each block is composed of convolutional layers, two batch normalizations, ReLU activation and residual skip connections. The block marked with the symbol $\triangledown$ has a dropout layer at its end. The block marked with 
the symbol $\triangleright$ is followed by a $1\times 1$ kernel convolutional layer with a number of features equal to the number of channels of the input image.}
\label{fig:AEAno}
\end{figure}

In this study, we considered different convolutional autoencoder architectures which vary in number of filters and size of the kernels. With respect to  the clustering approach,  these architectures leverage on the power of residual units and  are deeper networks. The residual 
blocks are connected through max pooling layers in order to decrease the resolution of the image. 
At the lowest level, a dropout layer is inserted to reduce overfitting and training time. The decoder part mirrors the encoder and is composed of three residual blocks, connected through upsampling layers in order to increase the resolution of the image back to its original size.
After the last decoder residual block, another convolutional layer with a $1\times 1$ kernel and a number of filter equals to the channels of the original image coupled with a logistic activation function, outputs the reconstruction of the input image. 
The considered networks differ from the number of filters in each block and/or kernel size. Namely, they are:
\begin{itemize}
\item Model S3: 2, 4, 8, 16, 8, 4, 2 filters and $3 \times 3$ kernels
\item Model S5: 2, 4, 8, 16, 8, 4, 2 filters and $5 \times 5$ kernels
\item Model M3: 4, 6, 8, 10, 8, 6, 4 filters and $3 \times 3$ kernels
\item Model M5: 4, 6, 8, 10, 8, 6, 4 filters and $5 \times 5$ kernels
\item Model B3: 32, 64, 128, 256, 128, 64, 32 filters and $3 \times 3$ kernels. This model is the 
one that in our studies performed best and we depict it in Fig~\ref{fig:AEAno}.
 We shall refer to it in the following also as the Ano-AE model. If not specified differently,
 the results presented below have been obtained using this architecture. 
\end{itemize}
The model is trained on healthy leaves, so we split the healthy samples in training, validation and test as follows:
20\% for testing, 10\% of the remaining samples for validation and the remainder for training. 
In this approach we also used data augmentation (translation, rotation, reflection, and zooming) and we 
obtained an enlarged dataset of~552 healthy samples.
We used He initialization and Adam optimizer with a starting learning rate $\eta = 10^{-3}$ and default hyperparameters. The maximum number of epochs has been set to 500 and we have implemented early stopping monitoring the validation loss with a patience parameter of 20 epochs. The batch size has been set to 4. 
After each epoch, the whole training set is shuffled. 
For this study, we used the R,G,B, and NIR channels.

\subsubsection*{Reconstruction results}

\begin{figure}[h]
\centering
\includegraphics[width=\linewidth]{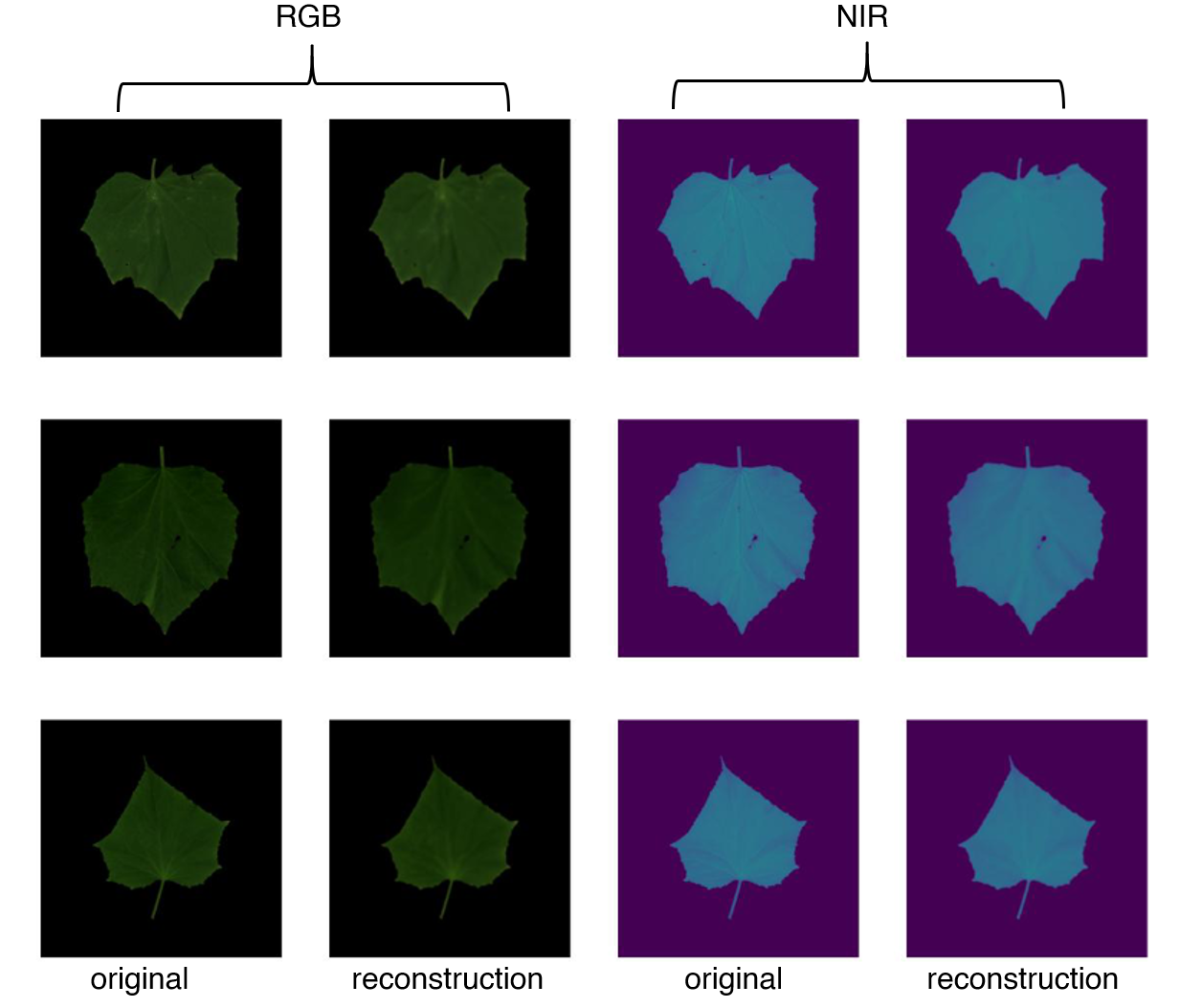}
\caption{{\bf  Comparison between original and reconstructed channels of healthy leaves belonging to the test set.}}
\label{fig:ANO_H}
\end{figure}

\begin{figure}[h]
\centering
\includegraphics[width=\linewidth]{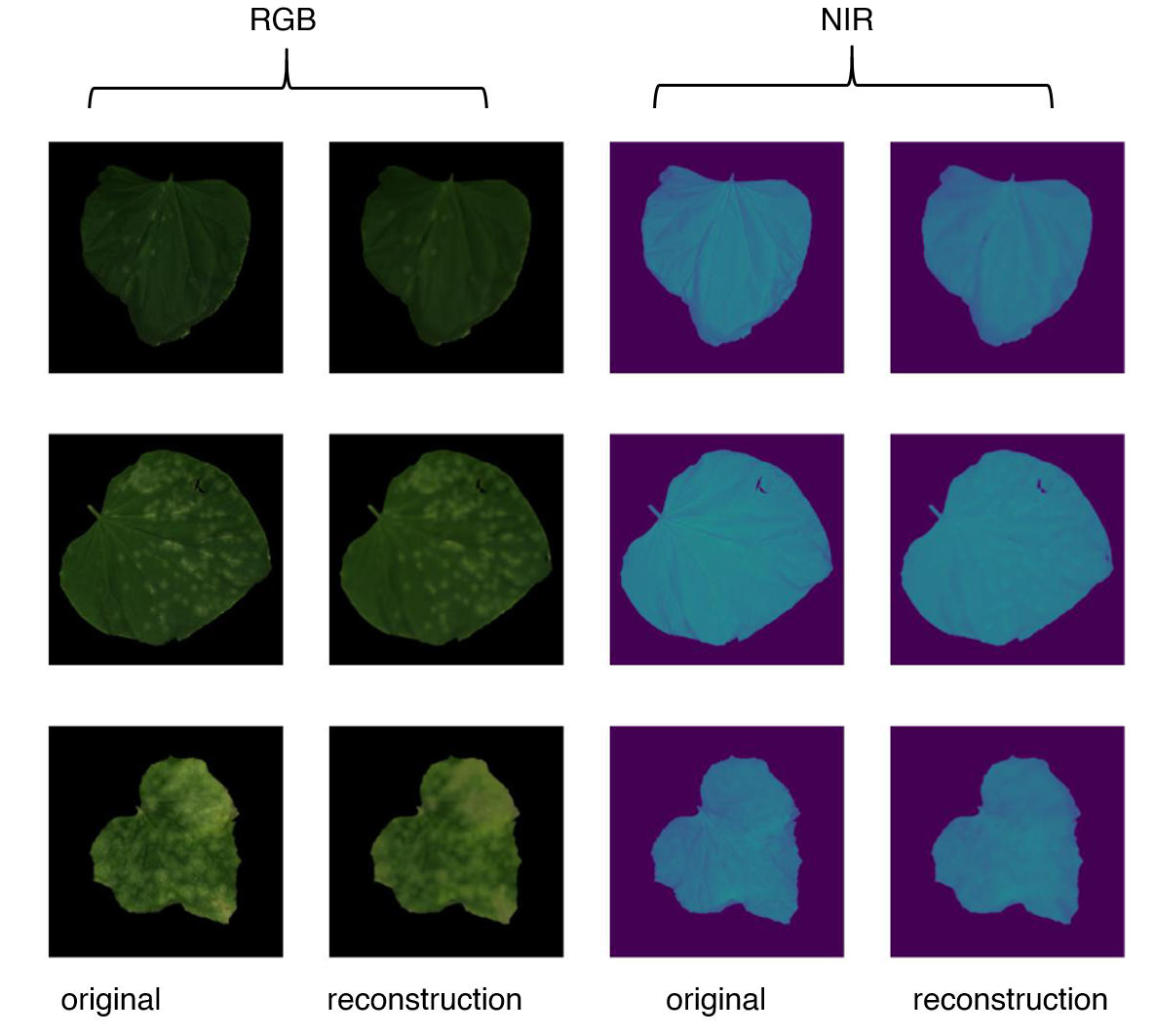}
\caption{{\bf  Comparison between original and reconstructed channels of diseased leaves belonging to the test set.}}
\label{fig:ANO_D}
\end{figure}

We start by analyzing the reconstructive power of the proposed architecture. 
First, we consider samples of healthy leaves (see Fig~\ref{fig:ANO_H}). 
The Ano-AE neural network is able to reconstruct the leaves, together with their characteristic traits like the stem, the veins, or different shades of color, with great accuracy. The reconstructed images show a certain blurriness compared to the original ones, like in the clustering approach, but the blur here is much less evident. 
Then, we check the reconstruction power on images of diseased leaves
(see Fig~\ref{fig:ANO_D}).  
It appears that the model can reconstruct the diseased leaves as well. 
However, the diseased spots are reconstructed in a more imprecise way, and with much more blur. Furthermore, the color of the disease spots appears to be different, being more brownish and slightly darker than the original one.
It is clear from the reconstruction of illness spots in diseased leaves that a number of healthy leaves in the training set must present some lesions or spots of other kinds (not disease--related) which have taught the model in the training phase  how to reconstruct also powdery mildew spots. 
For example, in the leaf of the first row of Fig~\ref{fig:ANO_D}, we observe the presence of a whitish mark 
which is similar to a powdery mildew spot. Of course, removing all healthy leaves which have such imperfections 
from the training set could increase the performance of the model, but this would make the model useless in 
a real-life scenario, where leaves are normally riddled with many different imperfections.

\subsubsection*{Anomaly detection results}

\begin{figure}[h]
\centering
\includegraphics[width=\linewidth]{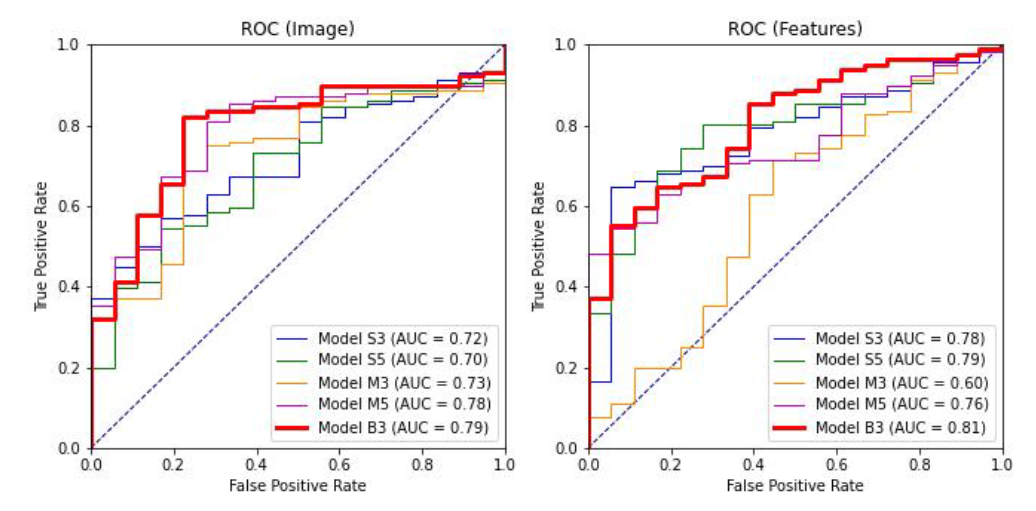}
\caption{{\bf  ROC curves and AUC values for the different neural architectures proposed for
anomaly detection.} The bold ROC corresponds to the best performing model (Ano-AE), the dashed curve corresponds
to random guess. The panel on the
left refers to the performance of the models when considering the image reconstruction error as anomaly score, while the panel on the right considers feature reconstruction error as anomaly score.}
\label{fig:ROC}
\end{figure}

We quantify the performance of the Ano-AE model using the Receiver Operating Characteristic (ROC) curve, which represents the probability of detection of false positives vs true positives by spanning all possible values of the threshold~$\Gamma$ (that is, each working point of a ROC curve is determined by a specific 
$\Gamma$ value).  Lowering the threshold more sample are classified as positive (anomalies), thus increasing both false positives and true positives. 
As compact measure of quality, we consider the area under the curve (AUC), which consists in the area below the ROC curve.   The AUC can be interpreted as the probability that a classifier gives a higher probability of being an anomaly to a randomly chosen abnormal (diseased) sample than to a randomly chosen normal (healthy) sample.
This, its measure ranges between 0 ({\em i.e.}, estimated labels are always wrong) and 1 ({\em i.e.}, estimated labels are always correct), passing through 0.5 ({\em i.e.}, random guess).
Fig~\ref{fig:ROC} shows the ROC curves obtained for the different neural architectures
introduced above for anomaly detection. Each curve shows the diagnostic ability of the model 
while varying the threshold~$\Gamma$ applied in the scoring system.
Results are reported both the image reconstruction error $s_x$ as 
anomaly score and the feature reconstruction error as anomaly score $s_z$.
As anticipated, it is apparent that model B3 (Ano-AE) performs better than the other investigated models, 
as its AUC is higher for both the scores.  

\section*{Conclusion}

Crop disease detection is a major challenge in crop protection and has a strong impact
on the subsequent management of diseased plants, both from
an economical viewpoint and an environmental/health--related perspective. 
Image--based detection strategies have the advantage of being capable to process
large canopies at an acceptable cost. 
In recent years, deep learning has opened a promising scenario, 
which allows the automated detection of the 
features relevant for disease detection starting from plant images. 
However DL nets do require a very ample set of training images in order to produce
accurate results. 
In our study, we have focused on unsupervised deep learning techniques
applied to multispectral imaging data for the detection of powdery mildew on cucumber leaves.
We have proposed the use of autoencoder architectures to obtain i) a clusterization of
the features, which by itself is not fully capable to provide accurate predictions but
do contain relevant information. In this respect, 
attention gates could be used in conjunction to feature clustering in order
to further filter the relevant ones;  
ii) an anomaly detection approach which has a significant potential of
resolution and that could be used as a prior for supervised nets
trained with a very limited number of labeled samples.

\section*{Author contribution}

{\bf Conceptualization:} Alessandro Benfenati, Paola Causin \\[2mm]
\noindent {\bf Dataset:} Alessandro Benfenati, Paola Causin, Roberto Oberti, Giovanni Stefanello \\[2mm]
\noindent
{\bf Methodology:} Alessandro Benfenati, Paola Causin, Giovanni Stefanello \\[2mm]
\noindent
{\bf Implementation:}  Giovanni Stefanello \\[2mm]
\noindent
{\bf Analysis:} Alessandro Benfenati, Paola Causin, Roberto Oberti, Giovanni Stefanello \\[2mm]
\noindent
{\bf Writing:} Alessandro Benfenati, Paola Causin, Roberto Oberti

\section*{Acknowledgments}
We acknowledge support from the SEED PRECISION
project (PRecision crop protection: deep learnIng and data fuSION), funded by 
Universit\`a degli Studi di Milano. AB also acknowledges the support of  
the GNCS group of INDAM (Istituto Nazionale di Alta Matematica "Francesco Severi").

\nolinenumbers

\bibliography{manuscript}

\begin{thebibliography}{10}

\bibitem{cisternas2020systematic}
Cisternas I, Vel{\'a}squez I, Caro A, Rodr{\'\i}guez A.
\newblock Systematic literature review of implementations of precision
  agriculture.
\newblock Computers and Electronics in Agriculture. 2020;176:105626.

\bibitem{martinelli2015advanced}
Martinelli F, Scalenghe R, Davino S, Panno S, Scuderi G, Ruisi P, et~al.
\newblock Advanced methods of plant disease detection. A review.
\newblock Agronomy for Sustainable Development. 2015;35(1):1--25.

\bibitem{gulhane2011detection}
Gulhane VA, Gurjar AA.
\newblock Detection of diseases on cotton leaves and its possible diagnosis.
\newblock International Journal of Image Processing (IJIP). 2011;5(5):590--598.

\bibitem{pixia2013recognition}
Pixia D, Xiangdong W, et~al.
\newblock Recognition of greenhouse cucumber disease based on image processing
  technology.
\newblock Open Journal of Applied Sciences. 2013;3(01):27--31.

\bibitem{delalieux2008near}
Delalieux S, Somers B, Hereijgers S, Verstraeten W, Keulemans W, Coppin P.
\newblock A near-infrared narrow-waveband ratio to determine Leaf Area Index in
  orchards.
\newblock Remote Sensing of Environment. 2008;112(10):3762--3772.

\bibitem{vigier2004narrowband}
Vigier BJ, Pattey E, Strachan IB.
\newblock Narrowband vegetation indexes and detection of disease damage in
  soybeans.
\newblock IEEE Geoscience and Remote Sensing Letters. 2004;1(4):255--259.

\bibitem{zhang2017leaf}
Zhang S, Wu X, You Z, Zhang L.
\newblock Leaf image based cucumber disease recognition using sparse
  representation classification.
\newblock Computers and electronics in agriculture. 2017;134:135--141.

\bibitem{mohanty2016using}
Mohanty SP, Hughes DP, Salath{\'e} M.
\newblock Using deep learning for image-based plant disease detection.
\newblock Front Plant Sci. 2016;7:1419.

\bibitem{sladojevic2016deep}
Sladojevic S, Arsenovic M, Anderla A, Culibrk D, Stefanovic D.
\newblock Deep neural networks based recognition of plant diseases by leaf
  image classification.
\newblock Comput Intell. 2016;2016.

\bibitem{fuentes2017robust}
Fuentes A, Yoon S, Kim SC, Park DS.
\newblock A robust deep-learning-based detector for real-time tomato plant
  diseases and pests recognition.
\newblock Sensors. 2017;17(9):2022.

\bibitem{wspanialy2020detection}
Wspanialy P, Moussa M.
\newblock A detection and severity estimation system for generic diseases of
  tomato greenhouse plants.
\newblock Computers and Electronics in Agriculture. 2020;178:105701.

\bibitem{lin2019deep}
Lin K, Gong L, Huang Y, Liu C, Pan J.
\newblock Deep learning-based segmentation and quantification of cucumber
  powdery mildew using convolutional neural network.
\newblock Front Plant Sci. 2019;10:155.

\bibitem{behmann2014detection}
Behmann J, Steinr{\"u}cken J, Pl{\"u}mer L.
\newblock Detection of early plant stress responses in hyperspectral images.
\newblock ISPRS Journal of Photogrammetry and Remote Sensing. 2014;93:98--111.

\bibitem{mahlein2016plant}
Mahlein AK.
\newblock Plant disease detection by imaging sensors--parallels and specific
  demands for precision agriculture and plant phenotyping.
\newblock Plant Dis. 2016;100(2):241--251.

\bibitem{lowe2017hyperspectral}
Lowe A, Harrison N, French AP.
\newblock Hyperspectral image analysis techniques for the detection and
  classification of the early onset of plant disease and stress.
\newblock Plant Methods. 2017;13(1):1--12.

\bibitem{saleem2019plant}
Saleem MH, Potgieter J, Arif KM.
\newblock Plant disease detection and classification by deep learning.
\newblock Plants. 2019;8(11):468.

\bibitem{west2003potential}
West JS, Bravo C, Oberti R, Lemaire D, Moshou D, McCartney HA.
\newblock The potential of optical canopy measurement for targeted control of
  field crop diseases.
\newblock Annual review of Phytopathology. 2003;41(1):593--614.

\bibitem{sankaran2010review}
Sankaran S, Mishra A, Ehsani R, Davis C.
\newblock A review of advanced techniques for detecting plant diseases.
\newblock Computers and electronics in agriculture. 2010;72(1):1--13.

\bibitem{piernik2021study}
Piernik M, Morzy T.
\newblock A study on using data clustering for feature extraction to improve
  the quality of classification.
\newblock Knowledge and Information Systems. 2021; p. 1--35.

\bibitem{ROUSSEEUW198753}
Rousseeuw PJ.
\newblock Silhouettes: A graphical aid to the interpretation and validation of
  cluster analysis.
\newblock Journal of Computational and Applied Mathematics. 1987;20:53--65.
\newblock doi:{https://doi.org/10.1016/0377-0427(87)90125-7}.

\bibitem{4766909}
Davies DL, Bouldin DW.
\newblock A Cluster Separation Measure.
\newblock IEEE Transactions on Pattern Analysis and Machine Intelligence.
  1979;PAMI-1(2):224--227.
\newblock doi:{10.1109/TPAMI.1979.4766909}.

\end{thebibliography}

%
%
%

\end{document}